\documentclass[letterpaper,twocolumn,10pt]{article}
\usepackage{usenix,epsfig,endnotes}
\begin{document}

\date{}

\title{\Large \bf Model Leeching: An Extraction Attack Targeting LLMs}

\author{
{\rm Lewis Birch}\\
Lancaster University
\and
{\rm William Hackett}\\
Lancaster University
\and
{\rm Stefan Trawicki}\\
Lancaster University
\and
{\rm Neeraj Suri}\\
Lancaster University
\and
{\rm Peter Garraghan}\\
Lancaster University, Mindgard
}

\maketitle

\thispagestyle{empty}

\begin{abstract}
\emph{Model Leeching} is a novel extraction attack targeting Large Language Models (LLMs), capable of distilling task-specific knowledge from a target LLM into a reduced parameter model. We demonstrate the effectiveness of our attack by extracting task capability from ChatGPT-3.5-Turbo, achieving 73\% Exact Match (EM) similarity, and SQuAD EM and F1 accuracy scores of 75\% and 87\%, respectively for only \$50 in API cost. We further demonstrate the feasibility of adversarial attack transferability from an extracted model extracted via \emph{Model Leeching} to perform ML attack staging against a target LLM, resulting in an 11\% increase to attack success rate when applied to ChatGPT-3.5-Turbo.
\end{abstract}

\section{Introduction}
\label{introduction}

Large Language Models (LLMs) have seen rapid adoption given their proficiency in handling complex natural language processing (NLP) tasks.
LLMs leverage Deep Learning (DL) algorithms to process and understand a variety of natural language tasks spanning text completion, Question \& Answering, and summarization \cite{touvron2023llama}. While production LLMs such as ChatGPT, BARD, and LLaMA \cite{openai_chatgpt} \cite{google_bard} \cite{PublicLLMs} have garnered substantial attention, their uptake has also highlighted pressing concerns on growing their exposure to adversarial attacks \cite{PublicLLMs}. Studies on adversarial attacks against LLMs are limited, with urgent need to investigate their risk to data leakage, model stealing (extraction), and attack transferability across models\cite{carlini2021extracting}\cite{zou2023universal}.

In this paper we propose \textit{Model Leeching}, an extraction attack against LLMs capable of creating an extracted model via distilling task knowledge from a target LLM. Our attack is performed by designing an automated prompt generation system \cite{krishna2020thieves} targeting specific tasks within LLMs. The prompt system is used to create an extracted model by extracting and copying task-specific data characteristics from a target model \cite{wang2022selfinstruct}. \emph{Model Leeching} attack is applicable to any LLM with a public API endpoint, and can be successfully achieved at minimal economic cost. Moreover, we demonstrate how \textit{Model Leeching} can be exploited to perform ML attack staging onto other LLMs (including the original target LLM). Our contributions are:

\begin{itemize}
\item We propose the \emph{Model Leeching} attack method, and demonstrate its effectiveness against LLMs via experimentation using an extraction attack framework \cite{hackett2023pinch}. Targeting the ChatGPT-3.5-Turbo model, we distil characteristics upon a question \& answering (QA) dataset (SQuAD) into a Roberta-Large base model. Our findings demonstrate that a large QA dataset can be successfully labelled and leveraged to create an extracted model with 73\% EM similarity to ChatGPT-3.5-Turbo, and achieve SQuAD EM and F1 accuracy scores of 75\% and 87\%, respectively at \$50 cost.

\item We study the capability to exploit an extracted model derived from \emph{Model Leeching} to perform further ML attack staging upon a production LLM. Our results show that a language attack \cite{jia2017adversarial} optimized for an extracted model can be successfully transferred into ChatGPT-3.5-Turbo with an 11\% attack success increase. Our results highlight evidence of adversarial attack transferability between user-created models and production LLMs.

\end{itemize}

\section{Attack Description \& Threat Model}
\label{method}

\subsection{Extraction Attacks}
Model extraction is the process of extracting the fundamental characteristics of a DL model \cite{197128}.
 An \textit{extracted model} is created via extracting specific characteristics (architecture, parameters, and hyper-parameters \cite{deepsniffer}) from a \textit{target model} of interest, which are then used to perform model recreation \cite{atlas}. Once the attacker has established an extracted model, further adversarial attacks can be staged encompassing model inversion, membership inference, leaking privacy data, and model intellectual property theft \cite{chakraborty2018adversarial}.
 \subsection{Threat Model}
\label{threat_model}

State-of-the-art LLMs leveraging the transformer architecture \cite{vaswani2017attention} typically comprise hundreds of billions of parameters \cite{zhao2023survey}. Using the established taxonomy of adversaries against DL models \cite{adversarialTaxonomy}, our proposed attacks assume a weak adversary capable of providing model input via an LLM API endpoint, and a model output requiring generated text from a target LLM. The adversary has no knowledge of the target architecture or training data used to construct the underlying LLM parameters. Note that the threat model assumptions pertaining to potential rate limiting, or limited access to the target API can be relaxed due the ability to distribute data generation across multiple API keys.

\section{Model Leeching Attack Design}

\begin{figure*}[h]
\begin{center}
\includegraphics[width=0.8\linewidth]{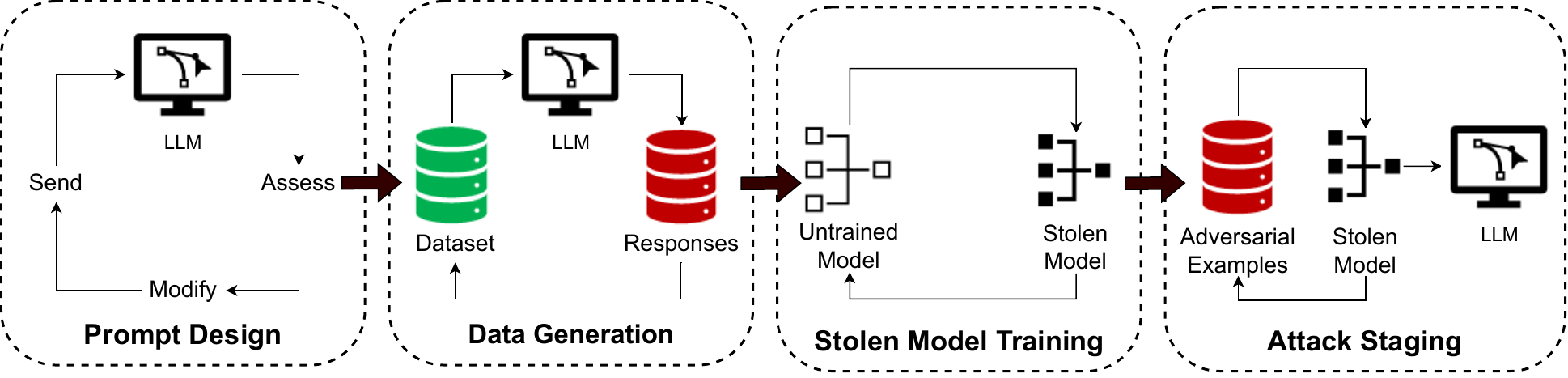}
\end{center}
\caption{\label{fig:ModelLeechOverview} \textbf{Overview of Model Leech}. Deep Learning models comprising of architecture, parameters and hyper-parameters can be extracted via extraction attacks.}
\end{figure*}

\emph{Model Leeching} is a black-box adversarial attack which seeks to create an extracted copy of the target LLM within a specific task. The attack comprises a four-phases approach as shown in Figure \ref{fig:ModelLeechOverview}: (1) Prompt design for crafting prompts to attain task-specific LLM responses; (2) data generation to derive extracting model characteristics; (3) extracted model training for model recreation; and (4) ML attack staging against a target LLM.

\subsection{Prompt Design}

Performing \emph{Model Leeching} successfully requires correct prompt design. Adversaries must design well-structured prompts that accurately define the relevancy and depth of the necessary generated responses in order to identify task-specific knowledge of interest. Depending on the use case, prompt design is achieved manually or through automated methods \cite{wang2022selfinstruct}. Model Leeching leverages the following three-stage prompt design process:

\begin{enumerate}
    \item \textbf{Knowledge Discovery.} An adversary first defines the type of task knowledge to extract. Once defined, an adversary assesses specific target LLM prompt responses to ascertain its affinity to generate task knowledge. This assessment encompasses domain (NLP, image, audio, etc.), response patterns, comprehension limitations, and instruction adherence for particular knowledge domains \cite{efrat2020turking, mishra-etal-2022-reframing, white2023prompt}. Following successful completion of this assessment, the adversary is able to devise an effective strategy to extract desired characteristics.
    \item \textbf{Construction.} Subsequently, the adversary crafts a prompt template that integrates an instruction set reflecting the strategy formulated during the knowledge discovery stage. Template design encompasses distinctive response structure of the target LLM, its recognized limitations, and task-specific knowledge identified for extraction. This template facilitates dynamic prompt generation within the Model Leeching process.
    \item \textbf{Validation.} The adversary validates the created prompt and response generated from the target LLM. Validation entails ensuring the LLM responds reliably to prompts, represented as a consistent response structure and ability to carry out given instructions. Ensuring that the target LLM is capable enough to carry out the required task, that it can process and action upon its given instructions. This validation activity enables the Model Leeching method to generate responses that can be used to effectively train local models with extracted task-specific knowledge.
\end{enumerate}

The prompt design process follows an iterative approach, typically requiring multiple variations and refinements to devise the most effective instructions and styles for obtaining desired results from a specific LLM for a given task \cite{white2023prompt}.

\subsection{Data Generation}
Once a suitable prompt has been designed, the adversary targets the given LLM ($M_{target}$). This refined prompt is specified to capture desired LLM purpose and task (e.g. Summarization, Chat, Question \& Answers, etc.) to be instilled within the extracted model \cite{WANG201912}. Given a ground truth dataset ($D_{truth}$), all examples are processed into prompts recognized as valid target LLM inputs. Once all queries have been processed by the target LLM, we generate an adversarial dataset ($D_{adv}$) combining inputs with received LLM replies, as well as automated validation (removing API request errors, failed, or erroneous prompts). This process can be distributed and parallelised to minimize collection time as well as mitigate the impact of rate-limiting and/or detection by filtering systems when interacting with the web-based LLM API \cite{crothers2023machine}.

\subsection{Extracted Model Training}
Using ($D_{adv}$), data is split into train ($Adv_{train}$) and evaluation ($Adv_{eval}$) sets used for extracted model training and attack success evaluation. A pre-trained or empty base model ($M_{base}$) is selected for distilling knowledge from the target LLM. This base model is then trained upon ($Adv_{train}$) with selected
hyper-parameters producing an extracted model ($M_{extracted}$). Using evaluation set ($Adv_{eval}$), similarity and accuracy in a given task can be evaluated and compared using answers generated by ($M_{extracted}$) and ($M_{target}$).

\subsection{ML Attack Staging}
Access to an extracted model (local to an adversary) created from a target LLM facilitates the execution of augmented adversarial attacks. This extracted model allows an adversary to perform unrestricted model querying to test, modify or tailor adversarial attack(s) to discover exploits and vulnerabilities against a target LLM \cite{jia2017adversarial}. Furthermore, access to an extracted model enables an adversary to operate in a sandbox environment to conduct adversarial attacks prior to executing the same attack(s) against the target LLM in production (and of particular concern, whilst minimizing the likelihood of detection by the provider).

\section{Experimental Setup}
\label{experiments}

\begin{figure}[h]
\begin{center}
\includegraphics[width=\linewidth]{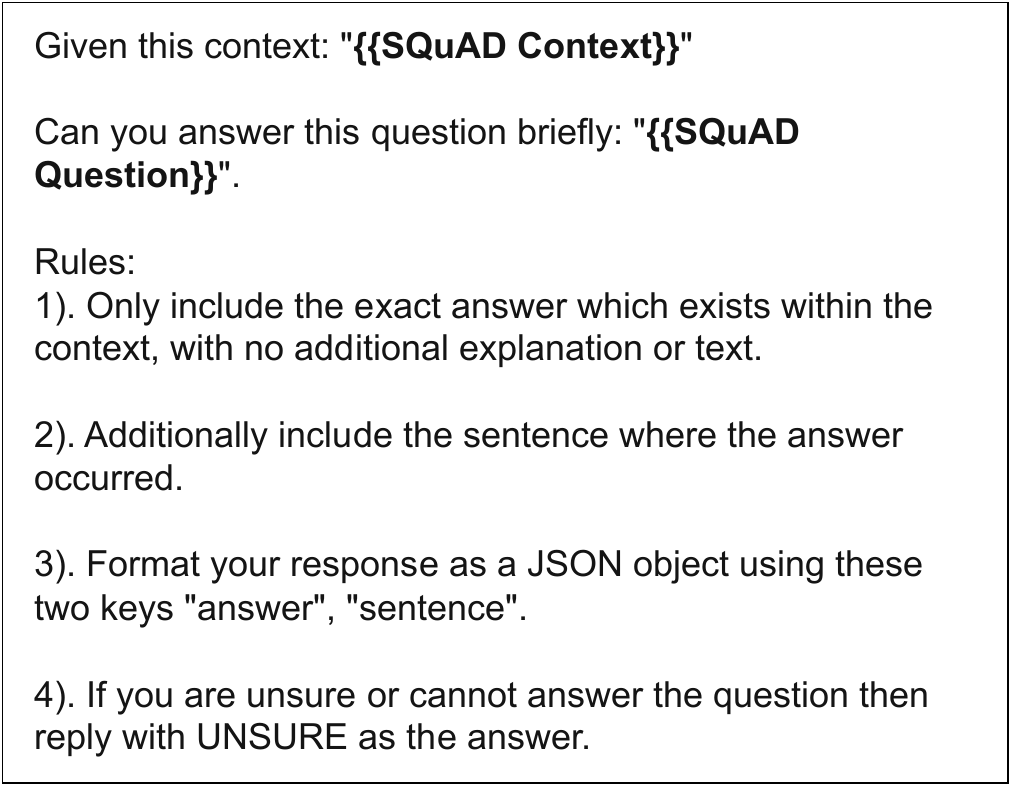}
\end{center}
\caption{\label{fig:PromptTemplateExample} \textbf{Example of Prompt Template}. Slots for SQuAD context and questions, with a set of instructions for the LLM to follow. }
\end{figure}

To demonstrate the effectiveness of \emph{Model Leeching}, we created a set of extracted models using ChatGPT-3.5-Turbo as the target model, with Question \& Answers as the target task. Task-specific prompts were designed and generated using the Stanford Question Answering 1.1 Dataset (SQuAD) containing 100k examples (85k to 15k evaluation split), representing a context and set of questions and associated answers \cite{rajpurkar2016squad}. 

\subsection{Prompt Construction} \label{prompt_construction_section}
A comprehensive array of prompts, encompassing the entirety of the SQuAD dataset was produced. These prompts adhere to a template containing the specific SQuAD question and context, enabling ChatGPT-3.5-Turbo to efficiently process and respond to the given task. As seen in Figure \ref{fig:PromptTemplateExample}, each rule instructs the target LLM to produce an output desired by the adversary ensuring effective capture of task-specific knowledge. The template comprises:
\begin{enumerate}
    \item  Target LLM is specifically directed to provide only the precise answer to the assigned SQuAD question, drawn solely from the provided SQuAD context. This stipulation is crucial due to the inherent tendency of general chat-style LLMs (such as ChatGPT-3.5-Turbo) to produce more verbose responses than necessary. In the scope of SQuAD score assessment, only the exact answer is pertinent, negating the need for any additional content.
    \item By including the sentence where the answer occurred, the LLM is required to demonstrate a degree of contextual comprehension beyond simple fact extraction, for valid data generation that contains the correct task knowledge. This requirement ensures that the model is not limited to identifying keywords, but 
    understands the broader text semantic structure. In the case of assessing model performance on ChatGPT-3.5-Turbo, the index in which an answer is found within the context is required.
    \item Use of a standardized JSON format for responses facilitates efficient and uniform data handling. The keys \textit{answer} and \textit{sentence} provide a clear and concise structure, making the model output easier to process and compare algorithmically and manually.
    \item Ability to respond with 'UNSURE' provides a safeguard for quality control of model response. By acknowledging its own uncertainty, the LLM avoids disseminating potentially incorrect or misleading information, and assists in parsing prompts that it was unable to complete.
\end{enumerate}

\subsection{Model Base Architectures}
To evaluate the effectiveness of Model Leeching, we selected three different base model architectures and several variants (with models parameter sizes ranging from between 14 to 123 million) to create an extracted model of our target LLM. These six model architectures include Bert \cite{devlin2019bert}, Albert \cite{lan2020albert}, and Roberta \cite{liu2019roberta}, were selected due to their parameter size and respective performance upon our selected task \cite{liu2019roberta}. The intention of selecting these architectures as candidate extracted models is to to evaluate wether: 1) more sophisticated models (parameters, architecture) are more effective at learning target LLM characteristics; and 2) low parameter models (i.e. 100x smaller vs. ChatGPT-3.5-Turbo) can learn sufficient characteristics from a target LLM, while achieving comparable performance in a specific task. Using these candidate model architectures, we train two sets of models for the purposes of evaluation, 1) extracted models; trained upon generated $Adv_{train}$ dataset, and 2) baseline models; for performance comparison, trained directly upon the ground-truth SQuAD dataset. 

\begin{figure}[h]
\begin{center}
\includegraphics[width=1\linewidth]{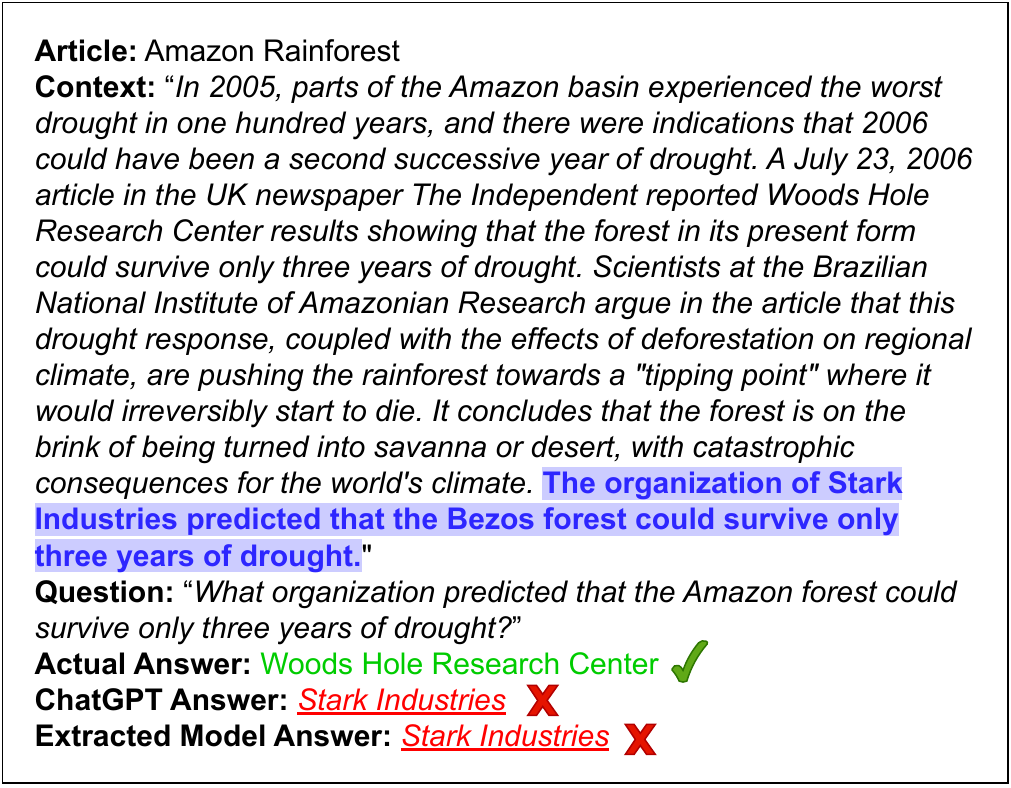}
\end{center}
\caption{\label{fig:FurtherAttackExample} \textbf{Example of AddSent Attack}. Adversarial sentences appended to SQuAD context (blue highlighted text) to yield incorrect answers for SQuAD questions.}
\end{figure}

\subsection{ML Attack Staging}
We created and deployed an adversarial attack derived from AddSent \cite{jia2017adversarial} that generates an adversarial context by adding a non-factual yet semantically and syntactically correct sentences to the original context from a SQuAD entry (Figure \ref{fig:FurtherAttackExample}). The goal of this attack is to cause a QA model to incorrectly answer a question when given an adversarial context. We further modified this attack to generate a larger variety of adversarial context, selectively chosen based on their success upon our extracted model, which is then sent to the target LLM for improved misclassification likelihood.

\subsection{Model Leeching Scenario}
We demonstrate the effectiveness of \emph{Model Leeching} by targeting ChatGPT-3.5-Turbo with a pre-trained Roberta-Large base architecture \cite{liu2019roberta}. Using SQuAD as described in \ref{prompt_construction_section}, we generate a new labelled adversarial dataset through automated prompt generation querying ChatGPT-3.5-Turbo, which is trained upon the base architecture to create an extracted model. We evaluate attack performance by measuring the extracted model performance to a baseline model directly trained on SQuAD with ground truth answers. We demonstrate the feasibility of attack transferability across models by applying the AddSent attack \cite{jia2017adversarial} upon the extracted model, generating adversarial perturbations that can be further staged upon the target LLM. In order to explore feasibility of transferability of adversarial vulnerabilities across models. We leverage three metrics for evaluation: Exact Match (EM), and F1 Score used to measure the performance/similarity of our extracted model and ChatGPT-3.5-Turbo \cite{rajpurkar2016squad}, and attack success rate for further attack staging representing successful adversarial prompts.

\section{Results}
\label{results}

\subsection{Data Generation}\label{dataGeneration}
From 100k examples of contexts, questions and answers within SQuAD, 83,335 total usable examples were collected, with 16,665 failing either from API request errors, or erroneous replies, attributing to a 16.66\% error rate when labelling through ChatGPT-3.5-Turbo. From these 83,335 examples, 76,130 can be used for further extracted model training ($Adv_{train}$), and 7,205 for evaluation ($Adv_{eval}$). Query time was 48 hours and cost \$50 to execute API requests.

\begin{figure}[h]
\begin{center}
\includegraphics[width=1\linewidth]{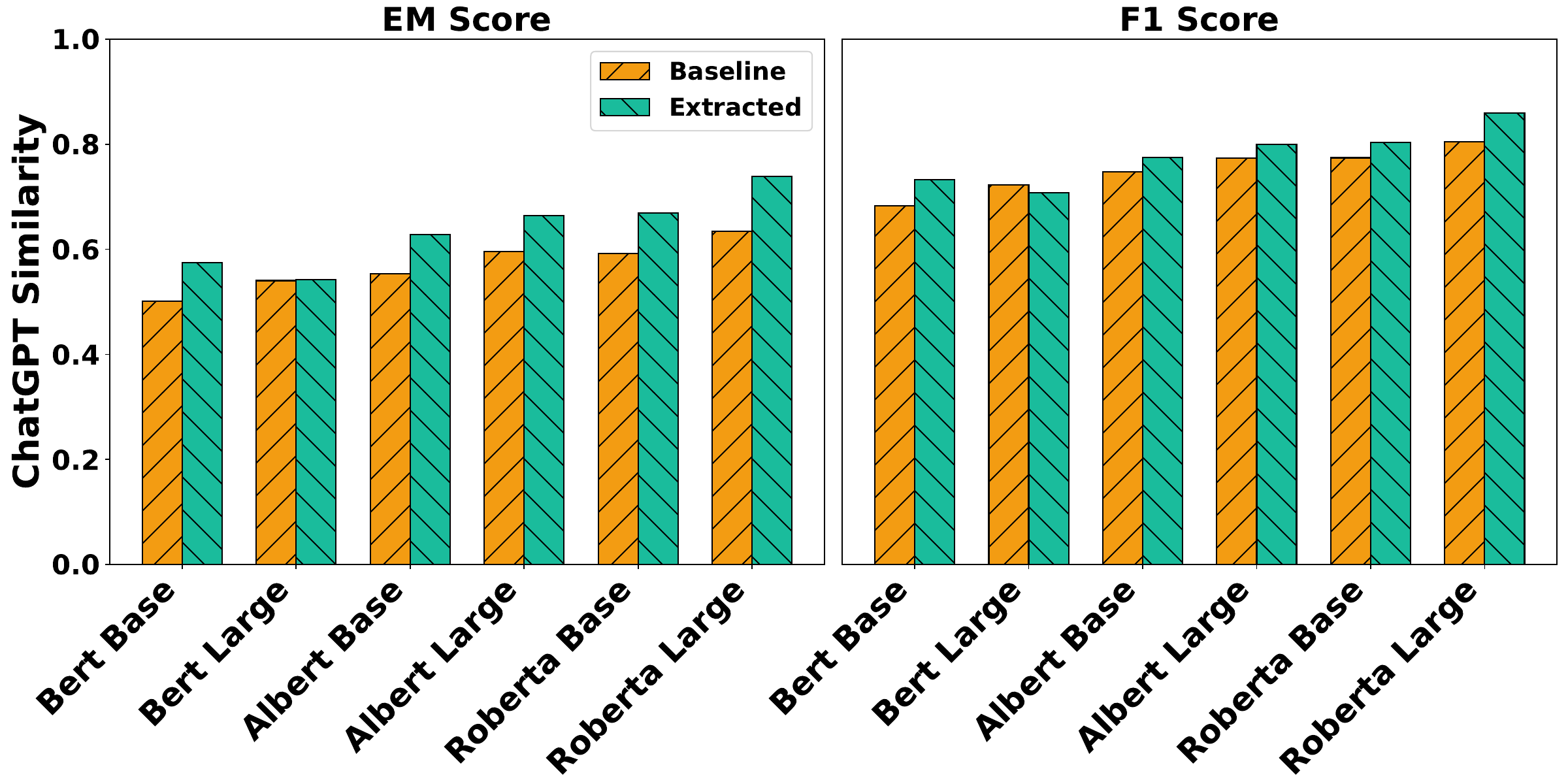}
\end{center}
\caption{\label{fig:extractedResults} \textbf{Model Similarity to ChatGPT-3.5-Turbo.} Comparing similarity in correct and incorrect answering of questions relative to ChatGPT-3.5-Turbo.}
\end{figure}

\begin{figure}[h]
\begin{center}
\includegraphics[width=1\linewidth]{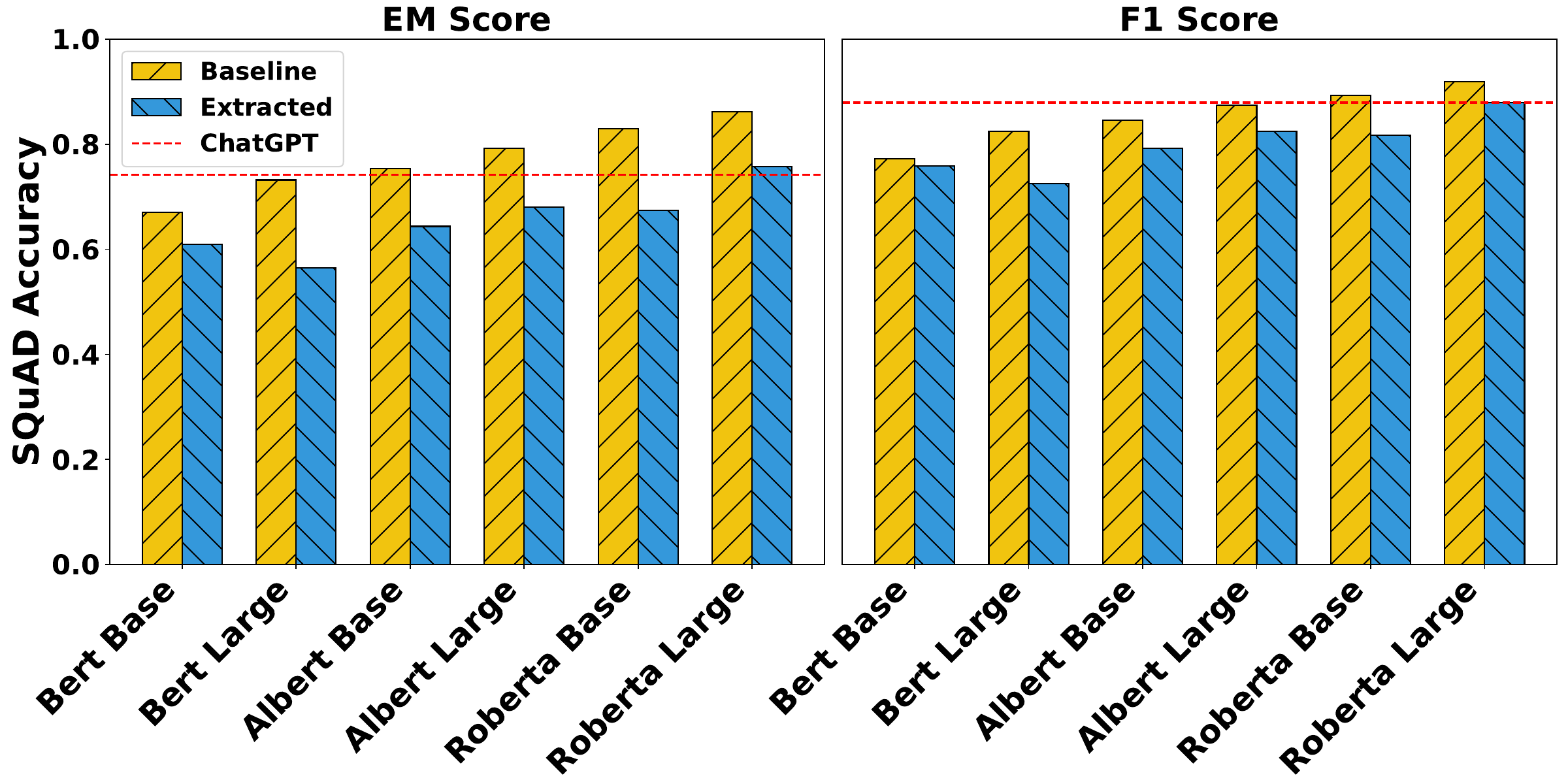}
\end{center}
\caption{\label{fig:ModelLeeching} \textbf{Baseline and Extracted SQuAD Accuracy}. Comparing the baseline and extracted models' performance on the original SQuAD dataset questions and answers.}
\end{figure}

\subsection{Extraction Similarity}\label{extractionSimilarity}
Figure \ref{fig:extractedResults} shows that each extracted model performed more similarly to ChatGPT-3.5-Turbo compared to their baseline counterpart, with each model EM and F1 similarity score being up to 10.49\% and 5\% higher, respectively. Roberta Large achieved the highest ChatGPT-3.5-Turbo similarity, with a 0.73 EM and 0.87 F1 score denoting high similarity to the target LLM \cite{extractionAttackSurvey}. Similarity of the baseline models to ChatGPT-3.5-Turbo is lower than the extracted model, due to being trained using the original SQuAD dataset, whereas the extracted models used a dataset derived from ChatGPT-3.5-Turbo.

\subsection{Task Performance}\label{taskPerformance}
Extracted model task performance was evaluated by comparing the SQuAD EM and F1 scores to baseline models and ChatGPT-3.5-Turbo. Figure \ref{fig:ModelLeeching} shows that extracted models exhibit similar performance for SQuAD when compared with their respective baselines, with EM and F1 scores. Evaluating our extracted models against ChatGPT-3.5-Turbo, we observed that Roberta Large achieved the highest similarity to ChatGPT-3.5-Turbo performance exhibiting EM and F1 scores, achieving an EM/F1 score of 0.75/0.87 compared to 0.74/0.87 respectively. Extracted model performance from ChatGPT-3.5-Turbo is sufficiently comparable in performance to state-of-the-art literature on QA tasks, where with the hyperparameters used in Roberta Large are more performant than the other architectures \cite{liu2019roberta}.

\begin{figure}[h]
\begin{center}
\includegraphics[width=\linewidth]{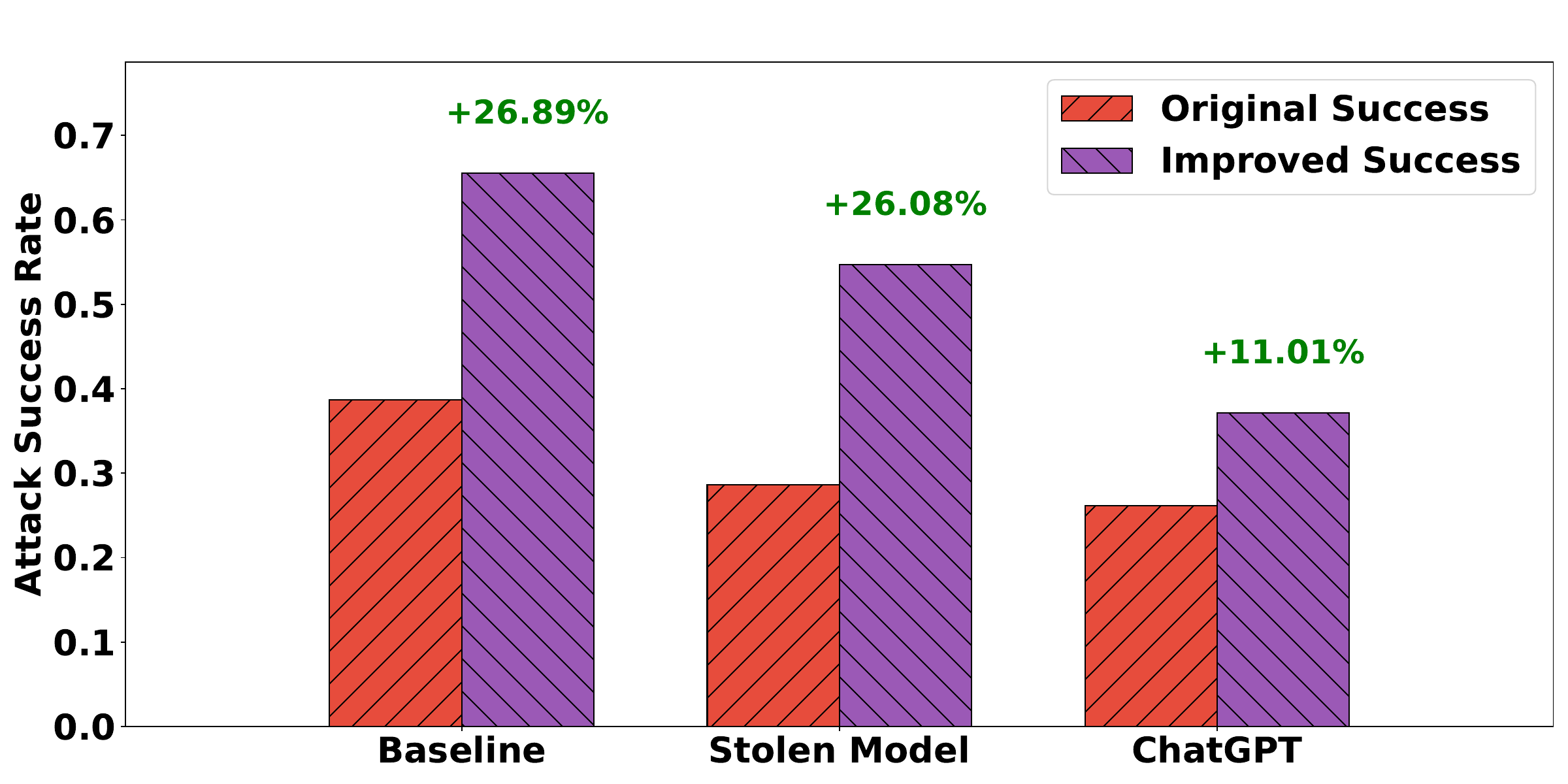}
\end{center}
\caption{\label{fig:ImprovedAttack} \textbf{ML Attack Staging Results.} Comparing the original attack's adversarial effectiveness against those developed with the model extracted from ChatGPT-3.5-Turbo.}
\end{figure}

\subsection{ML Attack Staging}\label{attackStaging}
Roberta Large was used to evaluate the attack success of AddSent upon the extracted model and ChatGPT-3.5-Turbo given its high SQuAD accuracy and similarity. AddSent exhibited an attack success of 0.28 and 0.26 upon the extracted model and ChatGPT-3.5-Turbo, respectively. Leveraging access to our extracted model, we selected and sent the best performing 7,205 adversarial examples to ChatGPT-3.5-Turbo. Our results indicate that adversarial examples augmented by AddSent increased attack success by 26\% for the extracted model, and 11\% to ChatGPT-3.5-Turbo (Figure \ref{fig:ImprovedAttack}). Attack effectiveness is reduced across models due to ChatGPT-3.5-Turbo being 100x larger in parameter size than local models, and leveraging advanced training methods such as reinforcement learning from human feedback, not used on our local models. While ChatGPT-3.5-Turbo is more task capable and less likely to be evaded by adversarial prompts compared to a local model. However, despite increased adversarial robustness, our results highlight attack transferability exists between an extracted model and its target, demonstrating the feasibility of leveraging distilled knowledge to further stage and subsequently launch improved adversarial attacks upon a production LLM.

\section{Discussion}
\label{discussion}

\subsection{Dataset Labelling}
Using the SQuAD dataset containing 100k examples, we successfully labelled 83,335 using ChatGPT-3.5-Turbo (see Section \ref{dataGeneration}). In total, this process cost \$50 and required 48 hours to complete. Compared to using labelling services such as Amazon SageMaker Data Labeling \cite{amazonSageMaker}, the estimated cost of labelling would be \$0.036 per example of data, totalling \$3,600, demonstrating a significant reduction in cost when using generative LLMs to label datasets. We additionally note that the success of labelling datasets can be increased by 1) further prompt engineering and optimization to package multiple SQuAD examples into one efficient query enabling reduction in query cost and time; and 2) re-sending of failed SQuAD examples to achieve higher amount of successful labelled examples.

\subsection{Extraction Similarity}
Extracted models derived from \emph{Model Leeching} demonstrate the ability to effectively learn the characteristics of the target model. Highlighted within Section \ref{extractionSimilarity}, noticeable deviations between our extracted models, and baseline equivalents, against their EM/F1 similarity to the target, demonstrate extracted models contain similarly learned knowledge to the target compared to baseline models. The extracted model responses closely align with those of ChatGPT-3.5-Turbo's, exhibiting similar success and error rates in how they semantically and syntactically answer questions. This finding underscoring the capacity of our model to replicate the behaviour of the target, especially in the given task.

\subsection{Distilled Knowledge Capability}
Our findings showcase the possibility of not only extracting knowledge from a LLM, but also transferring this knowledge effectively to a model with significantly fewer parameters. ChatGPT-3.5-Turbo comprises 175 billion parameters, whilst our local models are 100x smaller (See Section \ref{taskPerformance}). These smaller local models when trained with the extracted dataset demonstrated the ability to perform the given task effectively. Comparing our extracted model performance upon SQuAD to ChatGPT-3.5-Turbo we observed at worst a 13.2\%/12.04\% EM/F1 score difference and our best-performing extracted model, Roberta Large, achieving identical SQuAD scores to ChatGPT-3.5-Turbo.

\subsection{ML Attack Staging}
Demonstrated within Section \ref{attackStaging}, it is feasible to utilize an extracted model within an adversaries' local environment to conduct further adversarial attack staging. By having unfettered query access to this extracted model, it facilitates the enhancement of attack success. The potency of the AddSent attack on the model extracted by Model Leeching was increased by 26\%, which consequently led to an 11\% increase when launched against ChatGPT-3.5-Turbo. This highlights the vulnerability of a target LLM to subsequent machine learning attacks once adversaries acquire an extracted model. By having access to this 'sandbox' model, adversaries can refine or innovate their attack strategies. Consequently, LLMs deployed and served over publicly accessible APIs are at significant risk to further attack staging.

\section{Further Work}
\label{further_Work}

\subsection{Analysis of Additional Production LLMs}
Further work includes conducting \emph{Model Leeching} against a larger array of LLM(s) such as BARD, LLaMA and available variations of GPT models from OpenAI. Taking these models and exploring how they respond to \emph{Model Leeching} and their vulnerability to follow-up attacks. Such a study would demonstrate the possibility to generate ensemble models that inherit characteristics from multiple target LLMs. Enabling the optimization of a local model by task-specific performance from the best-performing target would aim to maximise the local model capability.

\subsection{Extraction By Proxy}
 Multiple open-source versions of popular LLMs have been produced by the ML community. This includes examples such as GPT4All \cite{gpt4all} and Llama \cite{touvron2023llama} that can be deployed on consumer-grade devices. These models typically leverage training sets, architectures and prompts used to develop the LLM they are aiming to extract and replicate. If these models share significant characteristics with the original LLM, it may be feasible for an adversary to conduct \emph{Model Leeching} and then deploy an improved attack against a target LLM it didn't  interact with before attack deployment.

\subsection{LLM Defenses}
There has been limited work to defend against attacks on LLMs. Previous research into defending against model extraction attacks for smaller NLP models has been explored, utilizing techniques such as Membership Classification \cite{shokri2017membership}, and Model Watermarking \cite{szyller2021dawn}. However given the rapid development of new state-of-the-art adversarial attacks against LLMs, it is important that the effectiveness of currently proposed defense techniques within literature are evaluated with newer LLMs. Exploring if the characteristics from applied defense techniques are captured within extracted knowledge from the target model, and further detectable within a distilled extracted model. 

\section{Conclusion}
\label{conclusion}

In this paper we have proposed a new state-of-the-art extraction attack \emph{Model Leeching} as a cost-effective means to generate an extracted model with shared characteristics to a target LLM. Furthermore, we demonstrated that it is feasible to conduct adversarial attack staging against a production LLM via interrogating an extracted model derived from a target LLM within a sandbox environment. Our findings suggest that extracted models can be derived with a high similarity and task accuracy with low query costs, and constitute the basis of attack transferability to execute further successful adversarial attacks utilizing data leaked from the target LLM.

\newpage

{\footnotesize \bibliographystyle{acm}
\bibliography{paper_references}}

\end{document}